\documentclass[sigconf,9pt]{acmart}
\setcopyright{none}
\renewcommand\footnotetextcopyrightpermission[1]{}
\AtBeginDocument{%
  \providecommand\BibTeX{{%
    \normalfont B\kern-0.5em{\scshape i\kern-0.25em b}\kern-0.8em\TeX}}}

\setcopyright{none}
\usepackage{subfigure}

\begin{document}
\title{Poster: Sponge ML Model Attacks of Mobile Apps \\
\vspace{3mm}
\large This work was published in ACM HotMobile 2023. Please cite DOI: 10.1145/3572864.3581586}

\author{Souvik Paul}
\affiliation{%
  \institution{University of Tartu}
    \country{Tartu, Estonia}}
 
\author{Nicolas Kourtellis}
\affiliation{%
  \institution{Telefonica Research}
    \country{Barcelona, Spain}}

\begin{CCSXML}
<ccs2012>
   <concept>
       <concept_id>10010147.10010178.10010219</concept_id>
       <concept_desc>Computing methodologies~Distributed artificial intelligence</concept_desc>
       <concept_significance>500</concept_significance>
       </concept>
   <concept>
       <concept_id>10002978.10003014.10003017</concept_id>
       <concept_desc>Security and privacy~Mobile and wireless security</concept_desc>
       <concept_significance>300</concept_significance>
       </concept>
 </ccs2012>
\end{CCSXML}

\ccsdesc[500]{Computing methodologies~Distributed artificial intelligence}
\ccsdesc[300]{Security and privacy~Mobile and wireless security}
\keywords{Adversarial Machine Learning, Latency
Attacks, Denial of Service}

\begin{abstract}
Machine Learning (ML)-powered apps are used in pervasive devices such as phones, tablets, smartwatches and IoT devices.
Recent advances in collaborative, distributed ML such as Federated Learning (FL) attempt to solve privacy concerns of users and data owners, and thus used by tech industry leaders such as Google, Facebook and Apple.
However, FL systems and models are still vulnerable to adversarial membership and attribute inferences and model poisoning attacks, especially in FL-as-a-Service ecosystems recently proposed~\cite{flaas2020}, which can enable attackers to access multiple ML-powered apps.
In this work, we focus on the recently proposed \textit{Sponge attack}: It is designed to soak up energy consumed while executing inference (not training) of ML model, without hampering the classifier's performance. 
Recent work~\cite{cina2022sponge} has shown sponge attacks on ASCI-enabled GPUs can potentially escalate the power consumption and inference time.
For the first time, in this work, we investigate this attack in the mobile setting and measure the effect it can have on ML models running inside apps on mobile devices.  
\end{abstract}

\maketitle
\pagestyle{plain}

\begin{figure}[t]
    \centering
    \includegraphics[width=1\linewidth]
    {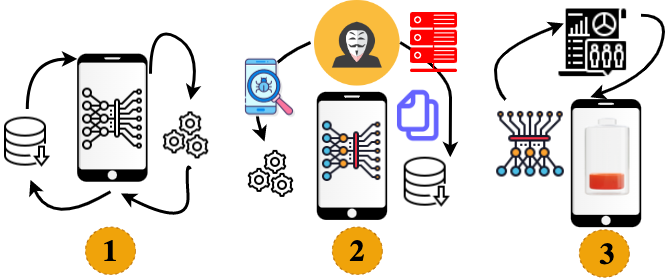}
    \caption{Sponge attack on mobile phones.}
    \label{fig:attackmodel}
\end{figure}

\section{Adversarial Sponge Model Attack}
Figure~\ref{fig:attackmodel} outlines the attack considered here:
(1) User/victim: operates an app that performs on-device training and generates a vanilla ML model.
(2) Attacker: a third-party app installed on the user's device that gained root system access and compromises the vanilla model built by the user app, by either installing a malicious patch that forces the victim's app to train a \textbf{sponge} model unknowingly, or replacing the final vanilla model with a \textbf{sponge} model trained on the attacker server.
(3) During inference phase, the installed sponge model generates proper classification results but also drains device battery.
We assume the attacker does not have access to the user's training data and only poisons the model.

\begin{figure}[!htb]
    \centering
    \includegraphics[width=1\linewidth]
    {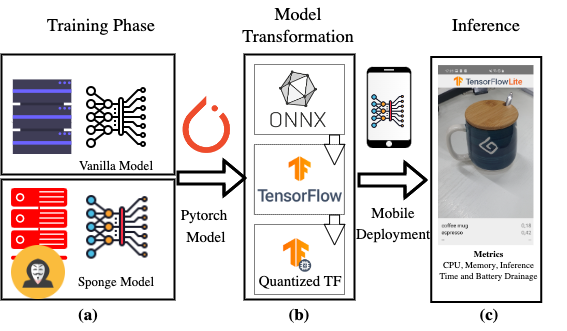}
    \caption{Experimental methodology.}
    \label{fig:method}
\end{figure}

\begin{figure*}[ht]
    \centering
        \begin{tabular}{cc}
        \includegraphics[width=0.4\linewidth]{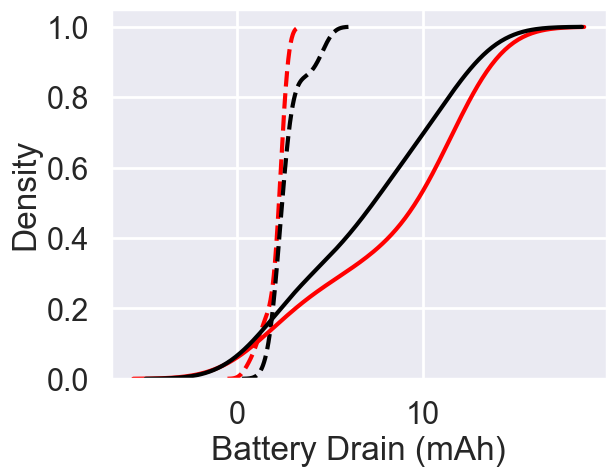}
         &
         \includegraphics[width=0.4\linewidth]{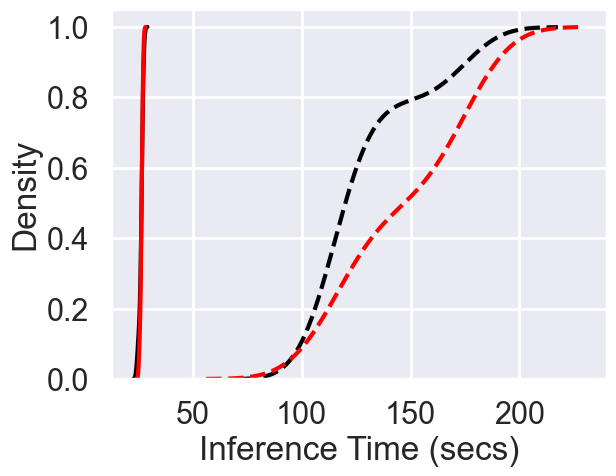}
          \\
          \includegraphics[width=0.4\linewidth]{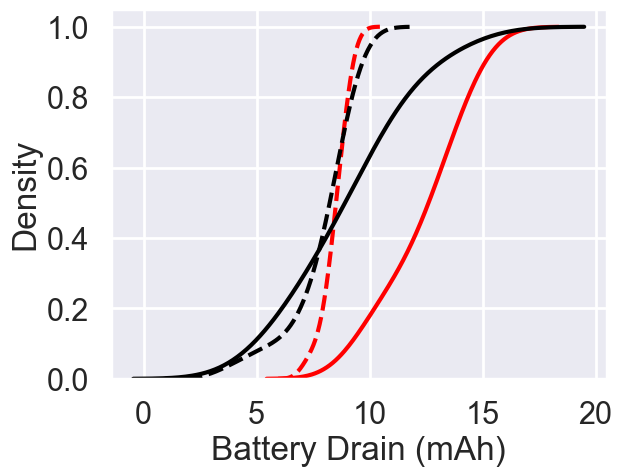}
          &
         \includegraphics[width=0.4\linewidth]{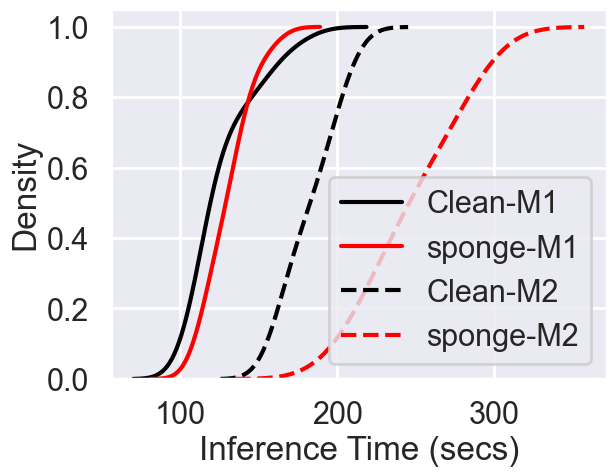}
        \end{tabular}
        % \vspace{-0.5cm}
        \caption{Preliminary experimental results (Samsung-S20 (top) and Nexus5 (below)).}
        \label{fig:preliminary-results}
        % \vspace{-0.5cm}
\end{figure*}

\section{Experimental Methodology}
Figure~\ref{fig:method} overviews the pipeline we follow to manifest the sponge attack on mobile devices.
(a) To keep model performance comparable, we train both vanilla and sponge ML models on a GPU-equiped server in PyTorch format for 100 epochs, following~\cite{cina2022sponge}.
We tune hyperparameters via grid search and find the best configuration for highest test accuracy and sponge effect.
(b) Then, we port each model to the victim's mobile by transforming the model from PyTorch to ONNX, then to Tensorflow, and then to Quantized Tensorflow Lite.
(c) Finally, we deploy each model into the mobile phone and perform automated inferences on at least 2000 test samples while measuring CPU and memory usage, total inference time and battery drainage.
We experiment with two models (MobileNetV2(M1) and ResNet18(M2)) on two mobile devices (Nexus 5 and Samsung-S20) using the FL benchmark dataset CIFAR-10.
We measure the first 3 metrics using the ADB shell and battery drainage through Android Battery Historian.
We repeat each setup 20 times, with the device fully charged and the display switched off during testing. 

\section{Results}
Figure~\ref{fig:preliminary-results} shows results for Samsung-S20~(top) and Nexus5~(below) for battery drain and inference time. For CPU and memory usage, we observe no significant difference during the inference phase for both devices, models M1 vs. M2, and sponge vs vanilla setting. \textbf{Key insights:}(1) Sponge attacks can be effective in increasing inference time by 13\% on average, and consequently draining the device battery faster by 15\%, on average.
(2) For high-end devices like Samsung S20 and optimized network MobileNetV2, there are more minor differences in inference time and battery drain; for low-end Nexus5, there is a marked increase in both metrics.
From these observations, we conclude that the Sponge attack is more effective when the model is deployed on low-end devices.

\section*{Acknowledgement}
This project received funding from the EU H2020 Research \& Innovation programme under grant agreements No 830927 (CONCORDIA) and 101021808 (SPATIAL). These results reflect only the authors’ view and the Commission is not responsible for any use that may be made of the information it contains.

\bibliographystyle{acm}
\bibliography{sample-base}

\end{document}